\documentclass[conference]{IEEEtran}
\IEEEoverridecommandlockouts
\usepackage{cite}
\usepackage{amsmath,amssymb,amsfonts}
\usepackage{algorithmic}
\usepackage{graphicx}
\usepackage{textcomp}
\usepackage{xcolor}

\newcommand{\Real}{\mathbb{R}}
\newcommand{\Natural}{\mathbb{N}}

\newcommand{\bfx}{\boldsymbol{x}}
\newcommand{\bfu}{\boldsymbol{u}}
\newcommand{\bfv}{\boldsymbol{v}}

\newcommand{\bfc}{\boldsymbol{c}}

\newcommand{\bfy}{\boldsymbol{y}}

\newcommand{\xivec}{\boldsymbol{\xi}}

\newtheorem{assume}{Assumption}
\newtheorem{problem}{Problem}
\newtheorem{thm}{Theorem}
\newtheorem{lem}{Lemma}
\newtheorem{rem}{Remark}

\def\BibTeX{{\rm B\kern-.05em{\sc i\kern-.025em b}\kern-.08em
    T\kern-.1667em\lower.7ex\hbox{E}\kern-.125emX}}

\begin{document}

\title{Learning from few examples\\ with nonlinear feature maps
}

\author{\IEEEauthorblockN{1\textsuperscript{st} Ivan Y. Tyukin}
\IEEEauthorblockA{\textit{Department of Mathematics} \\
\textit{King's College London}\\
London, UK \\
ivan.tyukin@kcl.ac.uk}
\and 
\IEEEauthorblockN{2\textsuperscript{nd} Oliver Sutton}
\IEEEauthorblockA{\textit{School of Computing }\\
\textit{ and Mathematical Sciences} \\
\textit{University of Leicester}\\
Leiceser, UK\\
oliver.sutton@leicester.ac.uk}
\and
\IEEEauthorblockN{3\textsuperscript{rd} Alexander N. Gorban}
\IEEEauthorblockA{\textit{ School of Computing  }\\
\textit{ and Mathematical Sciences} \\
\textit{University of Leicester}\\
Leiceser, UK\\
a.n.gorban@leicester.ac.uk}
}

\maketitle

\begin{abstract}
In this work we consider the problem of data classification in post-classical settings were the number of training examples consists of mere few data points.  We explore the phenomenon and reveal key relationships between dimensionality of AI model’s feature space, non-degeneracy of data distributions, and the model’s generalisation capabilities. The main thrust of our present analysis is on the influence of nonlinear feature transformations mapping original data into higher- and possibly infinite-dimensional spaces on the resulting model’s generalisation capabilities. Subject to appropriate assumptions, we establish new relationships between intrinsic dimensions of the transformed data and the probabilities to learn successfully from few presentations. 
\end{abstract}


\section*{Notation}

\begin{itemize}
	\item {$\Real$ denotes the field of real numbers, $\Real_{\geq 0}=\{x\in\Real| \ x\geq 0\}$, and} $\Real^n$ stands for the $n$-dimensional linear real vector space;
	\item $\Natural$ denotes the set of natural numbers;
	\item bold symbols $\boldsymbol{x} =(x_{1},\dots,x_{n})$ will denote elements of $\Real^n$;
	\item $(\boldsymbol{x},\boldsymbol{y})=\sum_{k} x_{k} y_{k}$ is the inner product of $\boldsymbol{x}$ and $\boldsymbol{y}$, and $\|\boldsymbol{x}\|=\sqrt{(\boldsymbol{x},\boldsymbol{x})}$ is the standard Euclidean norm  in $\Real^n$;
	\item  $\mathbb{B}_n$ denotes the unit ball in $\Real^n$ centered at the origin:
	\[\mathbb{B}_n=\{\boldsymbol{x}\in\Real^n| \ {\|\boldsymbol{x}\|\leq 1}\};\]
	\item  $\mathbb{B}_n(r,\bfy)$  stands for the ball in $\Real^n$ of radius ${r> 0}$ centered at $\bfy$: 
	\[\mathbb{B}_n(r,\bfy)=\{\boldsymbol{x}\in\Real^n| \ {\|\boldsymbol{x}-\bfy\|\leq r}\};\]
	\item $V_n$ is the $n$-dimensional Lebesgue measure, and $V_n(\mathbb{B}_n)$ is the volume of unit {$n$}-ball;
\end{itemize}

\section{Introduction}

Recent years have seen significant progress in the application of Artificial Intelligence (AI) and Machine Learning tools to a host of practically relevant tasks. Most importantly, we are witnessing major successes in the application of advanced large-scale models featuring millions of trainable parameters \cite{sandler2018mobilenetv2} to problems for which the volumes of available prior knowledge for training do not conform to the requirements of classical Vapnik-Chervonenkis theory \cite{vapnik1999overview} or other similar combinatorial bounds. A well-known example of the task in which this striking phenomenon can be observed is the MNIST digits dataset which, being reasonably small in size, can be learned remarkably well by modern large-scale deep neural networks. 

This property is fascinating in its own right, especially in view of  \cite{zhang2016understanding}, \cite{zhang2021understanding} reporting evidence that large-scale deep neural networks with identical architecture and training routines can both  successfully generalise beyond training data and at the same time overfit or memorise random noise. However, what is particularly striking is that some times an appropriately trained model is capable of exhibiting an extreme behaviour - learning from merely few presentations. 

To date, many different successful few-shot learning schemes have been reported in the literature. Matching \cite{vinyals2016matching} and prototypical \cite{snell2017prototypical} networks are examples of such learning machines. However, comprehensive theoretical justification of these schemes is yet to be seen. Recent work \cite{tyukin2021demystification}, \cite{gorban2021high} suggested a new framework offering a pathway for understanding of few-shot learning. Instead of focusing on classical ideas rooted in empirical risk minimisation coupled with distribution-agnostic bounds, it explores the interplay between the geometry of feature spaces and concentration of measure phenomena \cite{ledoux2001concentration}. This enables an escape from the apparent paradox of generalisation discovered in \cite{zhang2016understanding}, \cite{zhang2021understanding}.  

Instead of posing the question of generalisation for all possible data distributions, one can  ask a related but a different question: what properties of data distributions could be relevant or useful for few-shot learning? This refocusing might apparently be necessary in view of \cite{bartlett2020benign} showing that the spectrum of the data covariance matrix may hold the key to understanding benign overfitting.

In this work we adopt the theoretical framework proposed in \cite{tyukin2021demystification}, \cite{gorban2021high} and generalise it beyond the original setting whereby the problem of few-shot learning is analysed in models' native feature spaces. Here we explore how the problem of few-shot learning changes if one allows a nonlinear transformation of these features. Our motivation to study this question is two-fold. 

\begin{figure}
    \centering
    \includegraphics[width=0.475\textwidth]{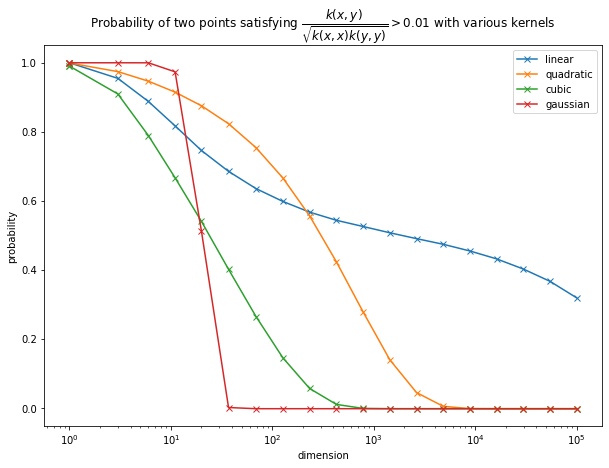}
    \includegraphics[width=0.475\textwidth]{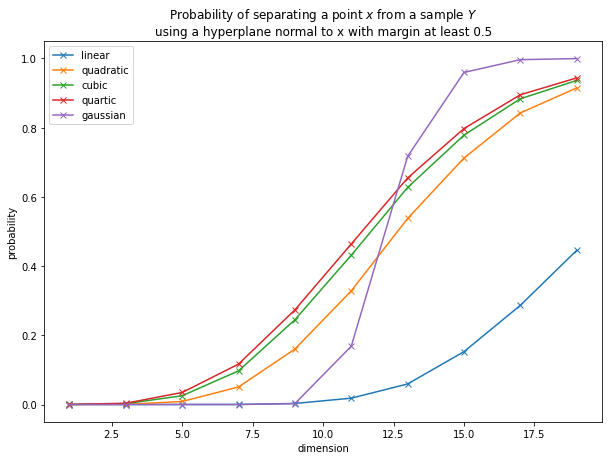}
    \caption{Empirical estimates of how easy it is to separate points using various nonlinear kernels. Top: separating points pairwise using kernel orthogonality. Bottom: separating a single point from a set of 20,000 other points using a linear separating surface in the kernel feature space. In both cases, the points are sampled from a uniform distribution in $[-1, 1]^n$. Here, $\phi_n(\bfx)$ denotes the image of the point $x \in \mathbb{R}^n$ under the kernel's associated feature mapping, and $\mu = |Y|^{-1} \sum_{y \in Y} \phi(y)$.}
    \label{fig:kernel_separability_orthogonality}
\end{figure}

First, many existing few-shot learning tools \cite{vinyals2016matching}, \cite{snell2017prototypical} already assume some sort of kernel-based transformation. Second, using kernels may enable mappings from original finite- or low-dimensional feature spaces into infinite-  or essentially high-dimensional spaces. The potential advantage of these transformations are illustrated in Fig. \ref{fig:kernel_separability_orthogonality}.
As these figures suggest, mapping vectors from their original spaces into their corresponding feature spaces induced by various kernels has a significant impact on data geometry in the mapped spaces. In particular, on the probability of the sample's quasi-orthogonality and linear separability. 

As we show here, the latter properties may offer new perspectives and capabilities affecting probabilities of success of such schemes. These results are stated formally in Theorem \ref{thm:few_shot} which is the main theoretical contribution of our work. 

The paper is organised as follows. In Section \ref{sec:preliminaries} we introduce some relevant notation and formulate the problem of few-shot learning, in which nonlinear feature transformations mapping input data into new feature spaces become important parameters of the problem. Section \ref{sec:main_results} presents  our main results including appropriate assumptions on the data distributions enabling the few-shot learning rules analysed in this work. These few-shot learning rules are very similar to those proposed and empirically studied in \cite{snell2017prototypical}. In this respect, Section \ref{sec:main_results} presents theoretical underpinnings for such rules. Section \ref{sec:conclusion} concludes the paper.


\section{Preliminaries and problem formulation}\label{sec:preliminaries}

In what follows we consider the problem of few-shot learning in the framework of a standard classification task. In this framework, we assume the existence of two sets of labels $\mathcal{L}$ and $\mathcal{L}_{new}$
\[
\mathcal{L}\cap\mathcal{L}_{new} = \emptyset,
\]
and two finite data sets,
\[\mathcal{X}=\{(\bfx,\ell) \  | \ \bfx\in\Real^n, \ \ell\in \mathcal{L}\}, \ |\mathcal{X}|=N,
\] 
and 
\[
\mathcal{Y}=\{(\bfx,\ell)  \ | \ \bfx\in\Real^n, \ \ell \in \mathcal{L}_{new}\}, \ |\mathcal{Y}|=k
\]
in which the pairs $(\bfx,\ell)\in\mathcal{X}$ are i.i.d. samples from some distribution $P_{\mathcal{X}}$, and the pairs $(\bfx,\ell) \in \mathcal{Y}$ are i.i.d. samples from some other distribution $P_{\mathcal{Y}}$. Elements $\ell\in\mathcal{L}\cup \mathcal{L}_{new}$ in the definitions of $\mathcal{X}$ and $\mathcal{Y}$ are the labels associated with the data vectors $\bfx$. 

In addition to the distributions $P_{\mathcal{X}}$ and $P_\mathcal{Y}$ it is convenient to consider the marginal distributions $P_{X}$ and $P_{Y}$:
\[
P_{X}(\bfx)=\sum_{\ell\in\mathcal{L}} P_{\mathcal{X}}(\bfx,\ell),
\]
\[
P_{Y}(\bfx)=\sum_{\ell\in\mathcal{L}_{new}} P_{\mathcal{Y}}(\bfx,\ell).
\]

We assume that there is a function $F$
\begin{equation}\label{eq:classifier_general}
F: \ \Real^n \rightarrow \mathcal{L}
\end{equation}
assigning an element from $\mathcal{L}$ to a vector from $\Real^n$. The function $F$ models expertise of the system in relation to it's capabilities to predict labels $\ell$ in the pairs $(\bfx,\ell)$ drawn from $\mathcal{Y}$ on the basis of the information that is contained in $\bfx$.

In this respect, the set $\mathcal{X}$ represents {\it existing knowledge} about the environment. This set may be arbitrarily large or even infinite, but the learner has no access to the elements from the set $\mathcal{X}$. The set $\mathcal{Y}$ represents {\it new knowledge} which is available to the learner. This new knowledge, however, is assumed to be scarce in the sense that $k\ll N$, $k \ll n$.

In addition to the data vectors $\bfx\in\Real^n$ we consider a parameterised family of feature maps $\phi_n$:
\begin{equation}\label{eq:phi_map}
\phi_n \ : \ \Real^n \rightarrow \mathbb{H}
\end{equation}
mapping elements of $\Real^n$ into a Hilbert space $\mathbb{H}$, which may be either finite- or infinite-dimensional. The map $\phi_n$ can represent transformations of the input data into the corresponding latent spaces in deep neural networks; it can also model other relevant data transformations emerging e.g. through the application of kernel tricks etc.

For every $\bfx\in\Real^n$, the map $\phi_n$, in turn, induces a kernel map $\kappa_n(\bfx,\cdot)$:
\[
\kappa_n(\bfx,\cdot): \ \Real^n \rightarrow \Real, \ \kappa_n(\bfx,\cdot)=(\phi_n(\bfx),\phi_n(\cdot)).
\]

\begin{rem} Examples of functions $\phi_n$ include the identity map $\phi_n(\bfx)=\bfx$ and feature maps of polynomial, $\kappa_n(\bfx,\bfy)=((\bfx,\bfy)+1)^m$, $m=1,2,\dots$, Gaussian $\kappa_n(\bfx,\bfy)=\exp(-\frac{\|\bfx-\bfy\|^2}{2\sigma^2})$, $\sigma\in\Real_{>0}$ and Laplacian $\kappa_n(\bfx,\bfy)=\exp(-\alpha \|\bfx-\bfy\|)$, $\alpha\in\Real_{>0}$ kernels.  
\end{rem}

The task is to learn a rule enabling the learner to discriminate between samples drawn from $P_{\mathcal{X}}$ and $P_{\mathcal{Y}}$ by accessing only the values of $\bfx_i$ and using available {\it training data} $\mathcal{Y}$, possibly some additional generic knowledge about $\mathcal{X}$, and the map $\phi_n$. More formally, the task is stated as follows (cf \cite{tyukin2021demystification}):

\begin{problem}[Few-shot learning]\label{prob:few_shot} Consider a classifier $F$ defined by (\ref{eq:classifier_general}), trained on a sample $\mathcal{X}$ drawn from some distribution $P_{\mathcal{X}}$. Let $\mathcal{Y}$ be a new sample that is drawn from another distribution $P_{\mathcal{Y}}$ and whose cardinality $|\mathcal{Y}|\ll n$.  Let $p_e,p_n\in(0,1]$ be given positive numbers determining the quality of learning. 

Find an algorithm $\mathcal{A}(\mathcal{Y})$ producing a new classification map 
\[
F_{new}: \mathcal{X}\rightarrow \mathcal{L}\cup \mathcal{L}_{new}
\]
such that
\begin{equation}\label{eq:learining_from_few_1}
P\big(F_{new}(\bfx) \in \mathcal{L}_{new} \big) \geq p_n
\end{equation}
for $\bfx$ drawn from $P_{Y}$, and
\begin{equation}\label{eq:learining_from_few_2}
P\big(F_{new}(\bfx) = F(\bfx)\big)\geq p_e
\end{equation}
for $\bfx$ drawn from the distribution $P_X$.
\end{problem}

\begin{rem}  Note that the set $\mathcal{L}_{new}$ in Problem \ref{prob:few_shot} is not necessarily a singleton. It may, in principle, contain more than one element. This allows questions to be posed regarding learning to discriminate between more than a single class.

The other point that is articulated in the statement of Problem \ref{prob:few_shot} is the requirement that $|\mathcal{Y}|\ll n$ defining the context of what ``few'' is referring to in the definition of few-shot learning problems.
\end{rem}
 
In the next section we describe sufficient conditions for the existence of algorithms $\mathcal{A}$ presenting a solution of the class of few-shot learning problems, as formulated in Problem \ref{prob:few_shot}.

\section{Main results}\label{sec:main_results}

We begin with the introduction of several useful characterisations of the maps $\phi_n$ in (\ref{eq:phi_map}) which will enable us to formulate appropriate requirements on the distributions $P_X$ and $P_Y$. Consider
\[
V_{\phi_n}(\bfc,r,n)=\int_{\|\phi_n(\bfx)-\bfc\|\leq r} 1 d\bfx.
\]
Symbol $n$ in the left-hand side of the above notation indicates that $\bfx$ are taken from $\Real^n$.

\begin{assume}\label{assume:rates}  There exists a function $\alpha_{\phi_n}: \mathbb{H}\times\mathbb{H}\times\mathbb{N}\rightarrow \Real_{\geq 0}$  such that for any $\bfc_1,\bfc_2\in\mathbb{H}$, $r_1\leq r_2\in\Real_{>0}$ the following holds true
\begin{equation}\label{eq:volume_rates}
\frac{V_{\phi_n}(\bfc_1,r_1,n)}{V_{\phi_n}(\bfc_2,r_2,n)} \leq C \left(\frac{r_1}{r_2}\right)^{\alpha_{\phi_n}(\bfc_1,\bfc_2,n)}
\end{equation}
\begin{equation}\label{eq:volume_rates}
V_{\phi_n}(\bfc_1,r_1,n) \leq f(r, n)
\end{equation}
whenever $V_{\phi_n}(\bfc_2,r_2,n)\neq 0$ and where the constant $C>0$ may be dependent on $\bfc_1$, $\bfc_2$.
\end{assume}

\begin{rem}
Note that the class of functions satisfying Assumption \ref{assume:rates} is not empty. It holds, for example, for $\phi_n(\bfx)= \bfx$ with $C=1$ and $\alpha_{\phi_n}(\bfc_1,\bfc_2,n)=n$.

In principle for some combinations of $\bfc_1,\bfc_2$ the constant $C$ may be infinite, although $C$ is guaranteed to be finite for $\bfc_1 = \bfc_2$ by the monotonic nature of $V_{\phi_n}$ whenever $V_{\phi_n}$ is finite. In what follows we will require that this constant exists and is finite for $\bfc_1,\bfc_2$ in a vicinity of some characteristic points in $\mathbb{H}$ determining concentration properties of data distributions (namely points $\bfc_X$ and $\bfc_Y$ in Assumptions \ref{assume:x}, \ref{assume:y} below).
We formalise this by supposing that
\[
C^{\ast}(\bfc,r)=\max_{\xivec: \ \|\bfc-\xivec\|\leq r }C(\xivec,\bfc),
\]
is finite for certain combinations of $\bfc$ and $r$. If the dependency of $C$ on $\bfc_1,\bfc_2$ is clear from the context then we will omit such explicit specifications in relevant expressions.
\end{rem}

For the functions $\alpha_{\phi_n}$  satisfying (\ref{eq:volume_rates}) we introduce
\begin{equation}\label{eq:beta}
\beta_{\phi_n}(\bfc,r,n)=\min_{\xivec: \ \|\bfc-\xivec\|\leq r} \alpha_{\phi_n}(\bfc,\xivec,n).
\end{equation}

We are now ready to proceed with specifying the requirements on $P_X$ and $P_Y$.

\begin{assume}\label{assume:x} For the distribution $P_X$, there is a corresponding probability density function $p_X$, positive numbers $A_X>0$, $r_X>0$,   and  $\bfc_X\in\mathbb{H}$, such that $p_X$ is supported on the set
\[
\mathcal{S}_X=\{\bfx\in\Real^n \ | \ \|\phi_n(\bfx)-\bfc_X\|\leq r_X\}, \ V_n(\mathcal{S}_X)>0,
\]
and satisfies the following growth bound:
\[
p_X(\bfx) \leq \frac{A_X}{V_{\phi_n}(\bfc_X,r_X,n)}.
\]
\end{assume}

\begin{assume}\label{assume:y} For the distribution $P_Y$, there is a corresponding probability density function $p_Y$, positive numbers $A_Y>0$, $r_Y>0$,  and  $\bfc_Y\in\mathbb{H}$, such that $p_Y$ is supported on the set
\[
\mathcal{S}_Y=\{\bfx\in\Real^n \ | \ \|\phi_n(\bfx)-\bfc_Y\|\leq r_Y\}, \ V_n(\mathcal{S}_Y)>0,
\]
and satisfies the following growth bound:
\[
p_Y(\bfx) \leq \frac{A_Y}{V_{\phi_n}(\bfc_Y,r_Y,n)} .
\]
\end{assume}

Observe that the functions $V_{\phi_n}$, $\beta_{\phi_n}$  in Assumptions \ref{assume:x}, \ref{assume:y} are determined exclusively by the feature maps $\phi_n$, whereas their arguments $\bfc_X$, $r_X$ and $\bfc_Y$, $r_Y$ capture relevant properties of $P_X$, $P_Y$. 
 
The rest of this Section is organised as follows. Our main result, Theorem \ref{thm:few_shot}, justifying solutions of the few-shot learning problem (Problem \ref{prob:few_shot}) with the help of some auxiliary functions
\[
\frac{1}{k}\sum_{i=1}^k \kappa_n(\bfx_i,\bfx) - \theta, \ \theta>0,
\]
where $\bfx_i$, $i=1,\dots,k$ are a part of the training sample, is stated and proved in Section \ref{sec:few_shot}. The proof of this theorem, however, is based on two other results. The first result is the generalised lemma on the typicality of quasi-orthogonality in high dimension (cf \cite{Kurkova}, \cite{kainen2020quasiorthogonal}, \cite{GorTyu:2016}) which we present in Section \ref{sec:sec:quasi_orthogonality}. The second result, which we call the law of high dimension, is presented in Section \ref{sec:sec:law_high_dimension}. Readers who may wish first to explore details of conditions and guarantees presented in our main theorem (Theorem \ref{thm:few_shot}) can skip the next two Sections and proceed to Section \ref{sec:few_shot}.
 
 \subsection{Quasi-orthogonality in Hilbert spaces}\label{sec:sec:quasi_orthogonality}
 
\begin{lem}[Quasi orthogonality]\label{lem:quasi_orthogonality}  Let $\mathcal{Z}=\{\bfx_1,\bfx_2,\dots,\bfx_k\}$ be a set of $k$ i.i.d. random vectors drawn from a distribution satisfying Assumption \ref{assume:y}, let $\delta,\varepsilon\in(0,1)$, and let $\phi_n$ satisfy Assumption \ref{assume:rates}.

Consider the event $A_1$:
\begin{equation}\label{eq:event_1}
A_1:  \ | (\phi_n(\bfx_i)-\bfc_Y,\phi_n(\bfx_j)-\bfc_Y)| \leq  {\delta r_Y}, \ \forall \  i\neq j
\end{equation}
and the event $A_2$:
\begin{equation}\label{eq:event_2}
A_2:  \  \|\phi_n(\bfx_i)-\bfc_Y\|\geq (1-\varepsilon)r_Y \ \forall \ i.
\end{equation}
Then 
\begin{equation}\label{eq:lem:orthogonality:statement:1}
P(  A_1 ) \geq 1 -  k (k-1) C A_Y \left[  (1-\delta^2)^{1/2}\right]^{\beta_{\phi_n}(\bfc_Y,r_Y\delta,n)},
\end{equation}
and
\begin{equation}\label{eq:lem:orthogonality:statement:2}
\begin{split}
&P\left(  A_1   \wedge  A_2 \right) \geq \\
& 1 - C A_Y k \left([1-\varepsilon]^{\beta(\bfc_Y,0,n)} \quad + \right.\\ 
& \quad \quad \ \left. (k-1)\left[(1-\delta^2)^{1/2}\right]^{\beta_{\phi_n}(\bfc_Y,r_Y\delta,n)}\right).
\end{split}
\end{equation}
\end{lem}

{\it Proof of Lemma  \ref{lem:quasi_orthogonality}}. Denote $\tilde{\phi}_i=\phi_n(\bfx_i)-\bfc_Y$ and consider the event
\[
E_1(\tilde{\phi}_1,\tilde{\phi}_2):  \ |(\tilde{\phi}_1/\|\tilde{\phi}_1\|,\tilde{\phi}_2)| > \delta.
\]
The probability that event $E_1(\tilde{\phi}_1,\tilde{\phi}_2)$ occurs is equal to 
\[
\int P(E_1(\tilde{\phi}_1,\tilde{\phi}_2) | \tilde{\phi_1}) p(\tilde{\phi}_1) d\phi_1.
\]
The conditional probability $P(E_1(\tilde{\phi}_1,\tilde{\phi}_2) | \tilde{\phi_1})$ is equal to the probability  that the vector $\tilde{\phi}_2$ ends up in the union of the following sets
\[
{\mathcal{C}_+}(\tilde{\phi}_1,\bfc_Y)=\left\{ \xivec \in \mathbb{H} \left|  \ \left(\frac{\tilde{\phi}_1}{\|\tilde{\phi}_1\|}, \xivec - \bfc_Y \right) > \delta   \right. \right\}
\]
\[
{\mathcal{C}_-}(\tilde{\phi}_1,\bfc_Y)=\left\{ \xivec \in \mathbb{H} \left|  \ \left(\frac{\tilde{\phi}_1}{\|\tilde{\phi}_1\|}, \xivec - \bfc_Y \right) < - \delta   \right. \right\}.
\]
Given that $\bfx_1,\dots,\bfx_k$ are drawn independently from the same distribution, this probability can be bounded from above as
\[
\begin{split}
&P(E_1(\tilde{\phi}_1,\tilde{\phi}_2)|\tilde{\phi}_1)=\int_{{\mathcal{C}_+}(\tilde{\phi}_1,\bfc_Y)} P_Y(\bfx)d\bfx\\
&\quad \quad \quad +\int_{{\mathcal{C}_-}(\tilde{\phi}_1,\bfc_Y)} P_Y(\bfx)d\bfx\\
&\leq \frac{A_Y}{V_{\phi_n}(\bfc_Y,r_Y,n)} \left(\int_{{\mathcal{C}_+}(\tilde{\phi}_1,\bfc_Y)} 1 d\bfx + \int_{{\mathcal{C}_-}(\tilde{\phi}_1,\bfc_Y)} 1 d\bfx \right).
\end{split}
\]
Observe that 
\[
\int_{{\mathcal{C}_+}(\tilde{\phi}_1,\bfc_Y)} 1 d\bfx < V_{\phi_n}(\bfc_+,r_Y (1-\delta^2)^{1/2},n)
\]
and 
\[
\int_{{\mathcal{C}_-}(\tilde{\phi}_1,\bfc_Y)} 1 d\bfx < V_{\phi_n}(\bfc_-,r_Y (1-\delta^2)^{1/2},n)
\]
for some $\bfc_+,\bfc_-\in\mathbb{H}$ satisfying
\[
\|\bfc_+-\bfc_Y\|\leq r_Y \delta, \ \|\bfc_- - \bfc_Y\|\leq r_Y\delta.
\]
Therefore, according to Assumption \ref{assume:rates} (eq. (\ref{eq:volume_rates}))
\[
\begin{split}
&P(E_1(\tilde{\phi}_1,\tilde{\phi}_2)|\tilde{\phi}_1)\leq C A_Y \left(  \left[(1-\delta^2)^{1/2}  \right]^{\alpha_{\phi_n}(\bfc_Y,\bfc_+,n)} \right.\\
&\left. + \left[(1-\delta^2)^{1/2}  \right]^{\alpha_{\phi_n}(\bfc_Y,\bfc_-,n)} \right).
\end{split}
\]
Taking (\ref{eq:beta}) into account, the above estimate results in
\begin{equation}\label{eq:bound_pair}
\begin{split}
& P(E_1(\tilde{\phi}_1,\tilde{\phi}_2)|\tilde{\phi}_1)\leq \\
& \quad 2 C A_Y  \left[(1-\delta^2)^{1/2} \right]^{\beta_{\phi_n}(\bfc_Y,r_Y\delta,n)}. \quad
\end{split}
\end{equation}
Hence, the probability that  the event $E_1(\tilde{\phi_1},\tilde{\phi_2})$ occurs admits the following upper bound:
\[
\begin{split}
&\int P(E_1(\tilde{\phi}_1,\tilde{\phi}_2)|\tilde{\phi}_1) p(\tilde{\phi}_1) d\tilde{\phi}_1 \leq \\
&2 C A_Y  \left[(1-\delta^2)^{1/2} \right]^{\beta_{\phi_n}(\bfc_Y,r_Y\delta,n)} \int  p(\tilde{\phi}_1) d\tilde{\phi}_1\\
&=2 C A_Y  \left[(1-\delta^2)^{1/2}  \right]^{\beta_{\phi_n}(\bfc_Y,r_Y\delta,n)}.
\end{split}
\]

Now consider events
\[
\begin{split}
& E_{m}(\tilde{\phi}_1,\dots,\tilde{\phi}_m):\\
& \left[\left|\left(\frac{\tilde{\phi}_1}{\|\tilde{\phi}_1\|},\tilde{\phi}_{m}\right)\right| > \delta \right] \vee \cdots \vee \left[\left|\left(\frac{\tilde{\phi}_{m-1}}{\|\tilde{\phi}_{m-1}\|}, \tilde{\phi}_m\right)\right| > \delta\right]
\end{split}
\]
for $m=2,\dots,k$. According to the union bound,
\[
\begin{split}
&P(E_{m}(\tilde{\phi}_1,\dots,\tilde{\phi}_m)|\tilde{\phi}_1,\dots,\tilde{\phi}_{m-1}) \leq\\
&\quad \quad \sum_{i=1}^{m-1} P\left(\left|\left(\frac{\tilde{\phi}_i}{\|\tilde{\phi}_i\|},\tilde{\phi}_{m}\right)\right| > \delta\left| \tilde{\phi}_i \right.\right)
\end{split}
\]
Applying the same argument as has been used in the derivation of (\ref{eq:bound_pair}), we can conclude that the right-hand side of the above inequality does not exceed the value of
\[
2 (m-1)  C A_Y  \left[(1-\delta^2)^{1/2}  \right]^{\beta_{\phi_n}(\bfc_Y,r_Y\delta,n)}.
\]
Hence
\begin{equation}\label{eq:bound_m_tuples}
\begin{split}
&P(E_{m}(\tilde{\phi}_1,\dots,\tilde{\phi}_m))\leq\\
&2 (m-1)  C A_Y  \left[(1-\delta^2)^{1/2}  \right]^{\beta_{\phi_n}(\bfc_Y,r_Y\delta,n)}
\end{split}
\end{equation}
for every $m=1,\dots,k$.

Now consider events
\[
B_m(\tilde{\phi}_m):  \ \|\tilde{\phi}_m\| < (1-\varepsilon)r_Y \ m=1,\dots,k.
\]
The probability $P(B_m(\tilde{\phi}_m)|\tilde{\phi_i},  \ i\neq m)$ is:
\begin{equation}\label{eq:norms_bound}
\begin{split}
& P(B_m(\tilde{\phi}_m)|\tilde{\phi_i},  \ i\neq m)=\int_{\|\phi_n(\bfx)-\bfc_Y\|\leq (1-\varepsilon)r_Y} P_Y(\bfx) d\bfx\\
& \leq  \frac{A_Y }{V_{\phi_n}(\bfc_Y,r_Y,n)} \int_{\|\phi_n(\bfx)-\bfc_Y\|\leq (1-\varepsilon)r_Y} 1 d\bfx\\
& = A_Y  \frac{V_{\phi_n}(\bfc_Y,(1-\varepsilon)r_Y,n)}{V_{\phi_n}(\bfc_Y,r_Y,n)} \\
& \leq C A_Y \left[(1-\varepsilon)\right]^{\beta_{\phi_n}(\bfc_Y,0,n)}
\end{split}
\end{equation}

Recall that for any events $\Omega_1,\dots,\Omega_d$ the following holds true:
\begin{equation}\label{eq:prob_bound}
P(\Omega_1 \land \Omega_2 \land \cdots \land \Omega_d)\geq 1 - \sum_{i=1}^d P(\mbox{not} \ \Omega_i).
\end{equation}
Therefore, using (\ref{eq:bound_m_tuples}) and (\ref{eq:norms_bound}), one can conclude that
\begin{equation}\label{eq:all_projections}
\begin{split}
& P((\mbox{not} \ E_1)\land \cdots \land(\mbox{not} \ E_k)) \geq 1 - \sum_{i=1}^d P(E_i)\\
& \geq 1 - k(k-1) C A_Y  \left[(1-\delta^2)^{1/2} \right]^{\beta_{\phi_n}(\bfc_Y,r_Y\delta,n)}
\end{split}
\end{equation}
and
\begin{equation}\label{eq:all_norms}
\begin{split}
& P((\mbox{not} \ B_1)\land \cdots \land(\mbox{not} \ B_k)) \geq 1 - \sum_{i=1}^d P(B_i)\\
& \geq 1 - k C A_Y  \left[(1-\varepsilon) \right]^{\beta_{\phi_n}(\bfc_Y,0,n)}.
\end{split}
\end{equation}

Finally, observe that $\|\tilde{\phi}_m\|$ is always bounded from above by $r_Y$. Therefore any $\tilde{\phi_1},\dots, \tilde{\phi_k}$  satisfying conditions
\[
\left[\left|\left(\frac{\tilde{\phi}_1}{\|\tilde{\phi}_1\|},\tilde{\phi}_{m}\right)\right| \leq \delta \right] \land \cdots \land \left[\left|\left(\frac{\tilde{\phi}_{m-1}}{\|\tilde{\phi}_{m-1}\|}, \tilde{\phi}_m\right)\right| \leq \delta\right]
\]
for $m=1,\dots,k$ must necessarily satisfy
\[
\left[\left|\left(\tilde{\phi}_1,\tilde{\phi}_{m}\right)\right| \leq \delta r_Y \right] \land \cdots \land \left[\left|\left(\tilde{\phi}_{m-1}, \tilde{\phi}_m\right)\right| \leq \delta r_Y\right].
\]

Hence, the event
$[\mbox{not} \ E_1 \wedge \cdots \wedge  \mbox{not} \ E_{k-1}]$ is contained in the event $A_1$ defined by (\ref{eq:event_1}) and 
\[
P(A_1)\geq   P(\mbox{not} \ E_1 \wedge\dots\wedge \mbox{not} \ E_{k-1}).
\]
and
\[
\begin{split}
& P(A_1 \wedge A_2)= P (A_1 \wedge \mbox{not}  \ B_1 \wedge \cdots \mbox{not}  \ B_k)\\ 
&\geq  P(\mbox{not} \ E_1 \wedge\dots\wedge \mbox{not} \ E_{k-1} \wedge  \mbox{not} \ B_1 \wedge \dots \wedge \mbox{not} \ B_k \cdots)\\
& \geq 1 - \sum_{i=1}^{k-1} P(E_i) - \sum_{i=1}^{k} P(B_i).
\end{split}
\]
This together with (\ref{eq:all_norms}), (\ref{eq:all_projections}) concludes the proof. $\square$

\subsection{The Law of High dimension in Hilbert Spaces}\label{sec:sec:law_high_dimension}
 
\begin{thm}[The law of high dimension]\label{thm:law_of_high_dimension} Consider a set $\mathcal{Z}=\{\bfx_1,\bfx_2,\dots,\bfx_k\}$ of $k$ i.i.d. random vectors drawn from a distribution satisfying Assumption \ref{assume:y}, and let the function $\phi_n$ satisfy Assumption \ref{assume:rates}.  Introduce the empirical mean of the sample in the feature space $\mathbb{H}$:
\[
\bar{\phi}_n  = \frac{1}{k} \sum_{i=1}^k \phi_{n}(\bfx_i).
\]
Finally, define 
\[
\begin{split}
U(k,\delta)=& k^{-1}(r_Y^2+(k-1)\delta r_Y), \\
L(k,\delta,\varepsilon) =& k^{-1}((1-\varepsilon)^2 r_Y^2 - (k-1)\delta r_Y),
\end{split}
\]
where $\delta,\varepsilon$ are some real numbers from $(0,1)$. 

Then the following holds for any $\delta,\varepsilon\in(0,1)$:
\begin{equation}\label{eq:lem:centering:2}
\begin{split}
 &P\left(\|\bar{\phi}_n-\bfc_Y\|^2 \leq U(k,\delta) \right) \geq\\
 & \quad \quad \quad \quad 1 - C A_Y k (k-1) \left[(1-\delta^2)^{1/2}\right]^{\beta_{\phi_n}(\bfc_Y,r_Y\delta,n)}.
 \end{split}
\end{equation}
Moreover, 
\begin{equation}\label{eq:lem:centering:1}
\begin{split}
& P\left(L(k,\delta,\varepsilon) \leq \|\bar{\phi}_n-\bfc_Y\|^2 \leq U(k,\delta) \right)\geq 1  \\
&  - \ C A_Y k[(1-\varepsilon)]^{\beta_{\phi_n}(\bfc_Y,0,n)} \\
& - \ C A_Y k (k-1) \left[(1-\delta^2)^{1/2}\right]^{\beta_{\phi_n}(\bfc_Y,r_Y\delta,n)}.
\end{split}
\end{equation}
\end{thm}
{\it Proof of Theorem \ref{thm:law_of_high_dimension}.}  
The proof follows from the Quasi-orthogonality Lemma (Lemma \ref{lem:quasi_orthogonality}). Consider 
\[
\begin{split}
& \|\bar{\phi}_n-\bfc_Y\|^2 = (\bar{\phi}_n-\bfc_Y, \bar{\phi}_n-\bfc_Y)\\
&= \left(\frac{1}{k}\sum_{i=1}^k \phi_n(\bfx_i) - \bfc_Y, \frac{1}{k}\sum_{i=1}^k \phi_n(\bfx_i) - \bfc_Y \right) \\
&=\frac{1}{k^2} \sum_{i=1}^k \|\phi_n(\bfx_i)-\bfc_Y\|^2 \\
& + \frac{1}{k^2} \sum_{i\neq j}  (\phi_n(\bfx_i) - \bfc_Y,\phi_n(\bfx_j) - \bfc_Y ).
\end{split}
\]
Lemma \ref{lem:quasi_orthogonality} (statement (\ref{eq:lem:orthogonality:statement:1})), states that the probability of that the below holds true
\[
 \frac{1}{k^2} \sum_{i\neq j}  \left|(\phi_n(\bfx_i) - \bfc_Y,\phi_n(\bfx_j) - \bfc_Y )\right| \leq \frac{k-1}{k} r_Y \delta
\]
is at least
\[
1 -  k (k-1) C A_Y \left[  (1-\delta^2)^{1/2}\right]^{\beta_{\phi_n}(\bfc_Y,r_Y\delta,n)}.
\]
Noticing that $\|\phi_n(\bfx_i)-\bfc_Y\|\leq r_Y $ for all $i=1,\dots,k$ assures that statement (\ref{eq:lem:centering:2}) holds. 

Combining the union bound, (\ref{eq:lem:centering:2}), and invoking statement  (\ref{eq:lem:orthogonality:statement:2}) of Lemma \ref{lem:quasi_orthogonality}, results in bound (\ref{eq:lem:centering:2}). $\square$

\subsection{Few-shot learning with nonlinear feature maps}\label{sec:few_shot}

\begin{thm}[Few-shot learning]\label{thm:few_shot} 
Let $F$ be a classifier defined by (\ref{eq:classifier_general}) and trained on a sample $\mathcal{X}$ drawn from some distribution $P_\mathcal{X}$ and whose marginal distribution $P_X$ satisfies Assumption \ref{assume:x} with $\bfc_X=0$. Let $\mathcal{Z}=\{\bfx_1,\dots,\bfx_k\}$,  $i=1,\dots,k$ be an i.i.d. sample drawn from a distribution $P_Y$  satisfying Assumption \ref{assume:y}, and whose corresponding class labels are from the set $\mathcal{L}_{new}$. Finally, suppose that the function $\phi_n$ satisfies Assumption \ref{assume:rates}.

Consider 
\[
D(\mathcal{Z})=\frac{1}{k} \left( \sum_{i=1}^{k} \sum_{j=1}^k \kappa_{n}(\bfx_i,\bfx_j) \right)^{1/2}
\]
and let $\delta\in(0,1)$ be a solution of
\[
\Delta=D(\mathcal{Z}) -\left(\frac{r_Y^2}{k}+\frac{k-1}{k} r_Y \delta\right)^{1/2} >0.
\]

Then the map
\begin{equation}\label{eq:learning_from_few_algorithm}
 F_{new}(\bfx)=\left\{\begin{array}{ll}
						   \ell_{\mathrm{new}}, & \frac{1}{k}\sum_{i=1}^k \kappa_n(\bfx_i, \bfx) - \theta D(\mathcal{Z}) \geq 0\\
						   F(\bfx) , & \mbox{otherwise}		
						   \end{array}\right.
\end{equation}
with $\ell_{new}\in\mathcal{L}_{new}$, parameterised by
\[
\theta\in \left[\max\{\Delta  - r_Y, 0\}, \Delta \right]
\]
is a solution of Problem \ref{prob:few_shot} with 
\begin{equation}\label{eq:thm:learning_few:bound_p_n}
\begin{split}
&p_n= \\
&\left(1 - C^\ast(\bfc_Y,\Delta-\theta) A_Y \times \right.\\
&\quad \quad \quad \left. \left[ \left(r_Y^2-(\Delta-\theta)^2\right)^{1/2}\right]^{\beta_{\phi_n}(\bfc_Y,\Delta-\theta,n)}\right) \times \\
 &\left(1- C^\ast(\bfc_Y,r_Y\delta) A_Y k (k-1) \times \right. \\ 
 & \quad \quad \quad \left. \left[ r_Y (1-\delta^2)^{1/2}\right]^{\beta_{\phi_n}(\bfc_Y,r_Y\delta,n)}\right),
\end{split}
\end{equation}
\begin{equation}\label{eq:thm:learning_few:bound_p_e}
p_e= 1 - C^\ast(0,\theta) A_X \left[\left(1-\frac{\theta^2}{r_X^2}\right)^{1/2}\right]^{\beta_{\phi_n}(0,\theta,n)}.
\end{equation}
\end{thm}
{\it Proof of Theorem \ref{thm:few_shot}}. The proof of the theorem relies on the law of high dimension property captured in Theorem \ref{thm:law_of_high_dimension}. According to this property,  the probability that the parameter $\bfc_Y\in\mathbb{H}$ determining concentration properties of the unknown distribution $P_Y$ is at most
\[
U(k,\delta)=\left(\frac{r_Y^2}{k}+\frac{k-1}{k}r_Y\delta\right)^{1/2}
\]
away in the space $\mathbb{H}$ from the empirical mean 
\[
\bar{\phi}_n=\sum_{i=1}^k \phi_n(\bfx_i)
\]
is at least
\begin{equation}\label{eq:prebound_p_n}
1- C^\ast(\bfc_Y,\delta r_Y) A_Y k (k-1)\left[(1-\delta^2)^{1/2}\right]^{\beta_{\phi_n}(\bfc_Y,\delta r_Y,n)}.
\end{equation}
Now, suppose that
\[
\|\bfc_Y-\bar{\phi_n}\|\leq U(k,\delta)
\]
holds true. Pick $0< \theta < \Delta$ and consider two sets:
\[
\mathcal{S}_1=\left\{\xivec\in\mathbb{H} \ | \ \left(\frac{\bar{\phi}_n}{\|\bar{\phi}_n\|},\xivec \right) - \theta = 0 \right\}
\]
and
\[
\mathcal{S}_2=\left\{\xivec\in\mathbb{H} \ | \ \left(\frac{\bar{\phi}_n}{\|\bar{\phi}_n\|},\xivec \right) - \left(\frac{\bar{\phi}_n}{\|\bar{\phi}_n\|},\bfc_Y \right) = 0 \right\}
\]
The sets $\mathcal{S}_1$ and $\mathcal{S}_2$ define hyperplanes in $\mathbb{H}$ which are parallel to each other with the set $\mathcal{S}_2$ containing the point $\bfc_Y$ (that is $\mathcal{S}_2$ passes through the vector $\bfc_Y$). We observe that
\[
\min_{\xivec: \ \|\xivec-\bar{\phi}_n\|\leq U(k,\delta)} \left(\frac{\bar{\phi}_n}{\|\bar{\phi}_n\|},\xivec\right)=\|\bar{\phi}_n\|-U(k,\delta) = \Delta,
\]
since $\|\bar{\phi}_n\|=D(\mathcal{Z})$,
and can therefore conclude that the set $\mathcal{S}_1$ is at least $D(\mathcal{Z})-U(k,\delta)-\theta=\Delta-\theta$ away from the set $\mathcal{S}_2$. 

Note that all points $\bfx\in\Real^n$ for which
\begin{equation}\label{eq:new_classifier_Y}
\begin{split}
& \left(\bar{\phi}_n, \phi_n(\bfx) \right) - \|\bar{\phi}_n\| \theta = \left(\bar{\phi}_n, \phi_n(\bfx) \right) - D(\mathcal{Z})\theta\\
& = \frac{1}{k}\sum_{i=1}^k \kappa_n(\bfx_i,\bfx)-D(\mathcal{Z})\theta > 0
\end{split}
\end{equation}
will be assigned label $\ell_{new}$ from $\mathcal{L}_{new}$ by the classifier $F_{new}$.

Let $\bfu$ be the orthogonal projection of $\bfc_Y$ onto the set $\mathcal{S}_1$. Then the probability that (\ref{eq:new_classifier_Y}) occurs for $\bfx$ drawn from $P_Y$ is 
\[
1-\int_{\mathcal{C}(\bfu,\|\bfu-\bfc_Y\|)} p_Y(\bfx)d\bfx,
\]
where 
\[
{\mathcal{C}(\bfu,d)}=\left\{ \bfx\in\Real^n \ \left| \left(\frac{\bfu-\bfc_Y}{\|\bfu-\bfc_Y\|},\phi_n(\bfx)-\bfc_Y\right)- d > 0 \right. \right\}.
\]
Noticing that $\|\bfu-\bfc_Y\|\geq \Delta-\theta$ since it is just the separation distance between $\mathcal{S}_1$ and $\mathcal{S}_2$, this probability is at least
\[
1-\int_{\mathcal{C}(\bfu,\Delta-\theta)} p_Y(\bfx)d\bfx.
\]
Taking Assumptions \ref{assume:rates}, \ref{assume:y}, the latter integral can be bounded from below as 
\[
1 - C^\ast(\bfc_Y,\Delta-\theta) A_Y \left[ \left(r_Y^2-(\Delta-\theta)^2\right)^{1/2}\right]^{\beta_{\phi_n}(\bfc_Y,\Delta-\theta,n)}.
\]
This together with (\ref{eq:prebound_p_n}) assures that (\ref{eq:thm:learning_few:bound_p_n}) holds.

Let $\bfx$ be drawn from $P_X$. The probability that $F_{new}(\bfx)\neq F(\bfx)$ is
\[
\int_{\left(\frac{\bar{\phi}_n}{\|\bar{\phi}_n\|},\phi_n(\bfx)\right)- \theta > 0} p_X(\bfx)d\bfx.
\]
Introducing $\bfv = \frac{\theta}{\| \bar{\phi}_n \|} \bar{\phi}_n$, this probability may be estimated by
\[
\int_{\|\bfv-\phi_n(\bfx)\|\leq (r_X^2 - \theta^2)^{1/2}} p_X(\bfx)d\bfx.
\]
\[
\begin{split}
&\int_{\|\bfv-\phi_n(\bfx)\|\leq (r_X^2 - \theta^2)^{1/2}} p_X(\bfx)d\bfx \\
& \leq  A_X \frac{1}{V_{\phi_n}(0,r_X,n)} \int_{\|\bfv-\phi_n(\bfx)\|\leq (r_X^2 - \theta^2)^{1/2}} 1 d\bfx\\
& =  A_X \frac{V_{\phi_n}(\bfv,(r_X^2 - \theta^2)^{1/2},n))}{V_{\phi_n}(0,r_X,n)} \\
&\leq C A_X \left[\left(1-\frac{\theta^2}{r_X^2}\right)^{1/2}\right]^{\alpha_{\phi_n}(\bfv,0,n)}\\
&\leq C^{\ast}(0, \theta) A_X \left[\left(1-\frac{\theta^2}{r_X^2}\right)^{1/2}\right]^{\beta_{\phi_n}(0,\theta,n)},
\end{split}
\]
and the bound (\ref{eq:thm:learning_few:bound_p_e}) follows. $\square$



\section{Conclusion}\label{sec:conclusion}

This paper provides, for the first time, a very general treatment of the challenge of few-shot learning. The main thrust of the work is to explicitly include the influence of non-linear feature transformations into the problem, assumptions, and solutions. The work determines key desired properties of these nonlinear transformations, captured by Assumption \ref{assume:rates}, as well as the properties of data, specified by Assumptions \ref{assume:x}, \ref{assume:y}, which are important for successful few-shot learning. 

These assumptions relate dimension of the original latent feature spaces with properties of nonlinear feature maps that are sufficient efficient learning. Potentially, these assumptions could also serve as explicit high-level specifications for the task of shaping or learning these nonlinear transformations from data. Detailed analysis of these properties and their practical feasibility are beyond the scope of this theoretical study. As our numerical examples show (see Fig. \ref{fig:kernel_separability_orthogonality}), exploration of the impact of nonlinear feature maps and their corresponding kernels on quasi-orthogonality, volume compression, and separability is a non-trivial and creative intellectual challenge which will be the focus of our future work.

\bibliographystyle{IEEEtran}
\bibliography{IEEEabrv,kernel_few_shot_refs}

\end{document}